\documentclass{article} % For LaTeX2e
\usepackage{iclr2022_conference,times}

\usepackage{amsmath,amsfonts,bm}

\def\eqref#1{equation~\ref{#1}}
\def\1{\bm{1}}

\def\mB{{\bm{B}}}

\def\mK{{\bm{K}}}

\def\mQ{{\bm{Q}}}

\def\mV{{\bm{V}}}

\DeclareMathAlphabet{\mathsfit}{\encodingdefault}{\sfdefault}{m}{sl}
\SetMathAlphabet{\mathsfit}{bold}{\encodingdefault}{\sfdefault}{bx}{n}

\usepackage{hyperref}
\usepackage{url}
\usepackage{amsmath}
\usepackage{amssymb}
\usepackage{multirow}
\usepackage{booktabs}
\usepackage{listings}
\usepackage{algorithm} 
\usepackage{graphicx}
\usepackage{color}
\usepackage{subcaption}
\usepackage{subcaption}
\usepackage{wrapfig}
 \usepackage{enumitem}

\title{CrossFormer:\,A\,Versatile\,Vision Transformer Hinging on Cross-scale Attention}

\author{Wenxiao Wang$^{1,2}$, Lu Yao$^1$, Long Chen$^3$, Binbin Lin$^4$, Deng Cai$^1$, Xiaofei He$^1$ \& Wei Liu$^2$ \\
    $^1$State Key Lab of CAD \& CG, Zhejiang University \\
    $^2$Data Platform, Tencent \\
    $^3$Columbia University \\
    $^4$School of Software Technology, Zhejiang University\\
}

\newcommand{\ie}{{\emph{i.e.}}}
\newcommand{\eg}{{\emph{e.g.}}}

 \iclrfinalcopy % Uncomment for camera-ready version, but NOT for submission.
\begin{document}

\maketitle
\vspace{-4mm}
\begin{abstract}
    Transformers have made great progress in dealing with computer vision tasks. However, existing vision transformers do not yet possess the ability of building the interactions among features of different scales, which is perceptually important to visual inputs. The reasons are two-fold: (1) Input embeddings of each layer are equal-scale, so no cross-scale feature can be extracted; (2) to lower the computational cost, some vision transformers merge adjacent embeddings inside the self-attention module, thus sacrificing small-scale (fine-grained) features of the embeddings and also disabling the cross-scale interactions. To this end, we propose \textbf{C}ross-scale \textbf{E}mbedding \textbf{L}ayer (CEL) and \textbf{L}ong \textbf{S}hort \textbf{D}istance \textbf{A}ttention (LSDA). On the one hand, CEL blends each embedding with multiple patches of different scales, providing the self-attention module itself with cross-scale features. On the other hand, LSDA splits the self-attention module into a short-distance one and a long-distance counterpart, which not only reduces the computational burden but also keeps both small-scale and large-scale features in the embeddings. Through the above two designs, we achieve cross-scale attention. Besides, we put forward a dynamic position bias for vision transformers to make the popular relative position bias apply to variable-sized images. Hinging on the cross-scale attention module, we construct a versatile vision architecture, dubbed CrossFormer, which accommodates variable-sized inputs. Extensive experiments show that CrossFormer outperforms the other vision transformers on image classification, object detection, instance segmentation, and semantic segmentation tasks.\footnote{The code has been released: \href{https://github.com/cheerss/CrossFormer}{https://github.com/cheerss/CrossFormer}}
%       The code has been released: \href{https://github.com/cheerss/CrossFormer}{https://github.com/cheerss/CrossFormer}.
\end{abstract}

 \vspace{-3mm}
\section{Introduction}
\vspace{-2mm}
It turns out that transformer~\citep{DBLP:conf/nips/VaswaniSPUJGKP17,DBLP:conf/naacl/DevlinCLT19,DBLP:conf/nips/BrownMRSKDNSSAA20} has achieved great success in the field of natural language processing (NLP). Benefitting from its self-attention module, transformer is born with the key ability to build long-distance dependencies. Since long-distance dependencies are also needed by a number of vision tasks~\citep{DBLP:journals/corr/abs-2105-13677,DBLP:journals/corr/abs-2104-13840}, a surge  of research work~\citep{DBLP:conf/iclr/DosovitskiyB0WZ21,DBLP:conf/icml/TouvronCDMSJ21,DBLP:journals/corr/abs-2102-12122} has been conducted to explore various transformer-based vision architectures.

A transformer requires a sequence of embeddings\footnote{In this paper, we also use ``embeddings'' to represent the input of each layer.}(\eg, word embeddings) as input. To adapt this requirement to typical vision tasks, most existing vision transformers~\citep{DBLP:conf/iclr/DosovitskiyB0WZ21,DBLP:conf/icml/TouvronCDMSJ21,DBLP:journals/corr/abs-2102-12122,DBLP:journals/corr/abs-2103-14030} produce embeddings by splitting an input image into equal-sized patches.
For example, a $224 \times 224$ image can be split into $56 \times 56$ patches of size $4 \times 4$, and these patches are projected through a linear layer to yield an embedding sequence.
Inside a certain transformer, self-attention is engaged to build the interactions between any two embeddings.
Thus, the computational or memory cost of the self-attention module is $O(N^2)$, where $N$ is the length of an embedding sequence. Such a cost is too big for a visual input because its embedding sequence is much longer than that of NLP. Therefore, the recently proposed vision transformers~\citep{DBLP:journals/corr/abs-2102-12122,DBLP:journals/corr/abs-2103-14030,DBLP:journals/corr/abs-2106-05786} develop multiple substitutes to approximate the vanilla self-attention module with a lower cost.

Though the aforementioned vision transformers have made some progress, they suffer from an issue that restricts their performance -- \textbf{They fail to build the interactions among features of different scales, whereas such an ability is very vital for a lot of vision tasks}. For example, an image often contains many objects of different scales, and to fully understand the image, building the interactions among those objects is helpful. Besides, some particular tasks such as instance segmentation need the interactions between large-scale (coarse-grained) features and small-scale (fine-grained) ones. Existing vision transformers fail to deal with the above cases due to two reasons:
(1) The embeddings are generated from equal-sized patches, so they only own features of one single scale. Moreover, their scales are kept unchanged or enlarged uniformly through operations like average pooling in the following layers. Hence, embeddings in the same layer are always equal-scale.
(2) Inside the self-attention module, adjacent embeddings are often grouped together and merged~\citep{DBLP:journals/corr/abs-2102-12122,DBLP:journals/corr/abs-2104-13840}. Since the number of groups is smaller than that of embeddings, such behavior can reduce the computational budget of the self-attention.
In this case, however, even if embeddings have both small-scale and large-scale features, merging operations will lose the small-scale (fine-grained) features of each individual embedding, thereby disabling the cross-scale attention.

To enable the building of cross-scale interactions, we co-design a novel embedding layer and self-attention module as follows. 1) \emph{Cross-scale Embedding Layer (CEL)} -- Following \citet{DBLP:journals/corr/abs-2102-12122}, we also employ a pyramid structure for our transformer, which naturally splits the vision transformer model into multiple stages. CEL appears at the start of each stage, which receives last stage's output (or an input image) as input and samples patches with multiple kernels of different scales (\eg, $4 \times 4$ or $8 \times 8$).
Then, each embedding is constructed by projecting and concatenating these patches as opposed to solely using one single-scale patch, which endows each embedding with cross-scale features.
2) \emph{Long Short Distance Attention (LSDA)} -- We propose a substitute of the vanilla self-attention, but to preserve small-scale features, the embeddings will not be merged. In contrast, we split the self-attention module into \textit{Short Distance Attention} (SDA) and \textit{Long Distance Attention} (LDA).
SDA builds the dependencies among neighboring embeddings, while LDA takes charge of the dependencies among embeddings far away from each other.
The proposed LSDA can also reduce the cost of the self-attention module like previous studies~\citep{DBLP:journals/corr/abs-2102-12122,DBLP:journals/corr/abs-2104-13840}, but different from them, LSDA does not undermine either small-scale or large-scale features. As a consequence, attention with cross-scale interactions is enabled.

Besides, following prior work~\citep{DBLP:conf/naacl/ShawUV18,DBLP:journals/corr/abs-2103-14030}, we employ a relative position bias for embeddings' position representations.
The Relative Position Bias (RPB) only supports fixed image/group size\footnote{Some vision transformers split input embeddings into several groups. Group size means the number of embeddings in a group.}. However, image size for many vision tasks such as object detection is variable, so does group size for many architectures, including ours.
To make the RPB more flexible, we further introduce a trainable module called \textit{Dynamic Position Bias} (DPB), which receives two embeddings' relative distance as input and outputs their position bias. The DPB module is optimized end-to-end in the training phase, inducing an ignorable cost but\,making\,RPB\,apply to variable image/group size.

All our proposed modules can be implemented with about ten lines of code.
Based on them, we construct four versatile vision transformers of different sizes, dubbed \emph{CrossFormers}. Other than image classification, the proposed CrossFormer can handle a variety of tasks with variable-sized inputs such as object detection. Experiments on four representative vision tasks (\ie, image classification, object detection, instance segmentation, and semantic segmentation) demonstrate that CrossFormer outperforms the other state-of-the-art vision transformers on all the tasks.
Remarkably, the performance gains brought by CrossFormer are substantially significant on dense prediction tasks, \eg, object detection and instance/semantic segmentation.

It is worth highlighting our contributions as follows:
\begin{itemize}[leftmargin=5mm]
    \vspace{-2mm}
    \item We propose cross-scale embedding layer (CEL) and long short distance attention (LSDA), which together compensate for existing transformers' incapability of building cross-scale attention.
%   \vspace{-1mm}
    \item The dynamic position bias module (DPB) is further proposed to make the relative position bias more flexible, \ie, accommodating variable image size or group size.
%   \vspace{-1mm}
    \item Multiple CrossFormers with different sizes are constructed, and we corroborate their effectiveness through sufficient experiments on four representative vision tasks.
%   \vspace{-2mm}
\end{itemize}

\vspace{-3mm}
\section{Background}
\vspace{-2mm}

\textbf{Vision Transformers.} Motivated by the transformers developed for NLP, researchers design specific visual transformers for vision tasks to take full advantage of their powerful attention mechanism. In particular, ViT and DeiT transfer the original transformer~\cite{DBLP:conf/nips/VaswaniSPUJGKP17} to vision tasks~\citep{DBLP:conf/icml/TouvronCDMSJ21,DBLP:conf/iclr/DosovitskiyB0WZ21}, achieving impressive performance. Later, PVT~\citep{DBLP:journals/corr/abs-2102-12122}, HVT~\citep{DBLP:journals/corr/abs-2103-10619}, Swin~\citep{DBLP:journals/corr/abs-2103-14030}, and ViTAE~\citep{DBLP:journals/corr/abs-2106-03348} introduce a pyramid structure into the visual transformers, greatly decreasing the number of patches in the later layers of a respective model. They also extend the visual transformers to other vision tasks like object detection and segmentation~\citep{DBLP:journals/corr/abs-2102-12122,DBLP:journals/corr/abs-2103-14030}.

\begin{figure*}[tb]
    \centering
    \includegraphics[width=1.0\linewidth]{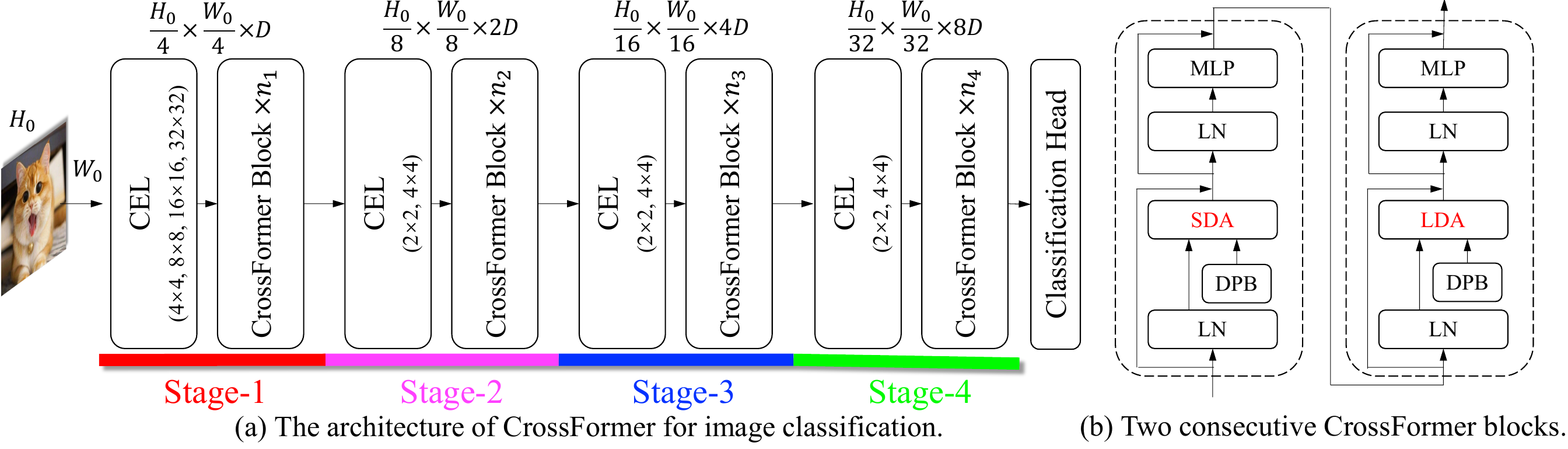}
    \caption{(a) The architecture of CrossFormer for classification. The input size is $H_0 \times W_0$, and the size of feature maps in each stage is shown on the top. \textit{Stage-i} consists of a CEL and $n_i$ CrossFormer blocks. Numbers in CELs represent kernels' sizes used for sampling patches. (b) The inner structure of two consecutive CrossFormer blocks. SDA and LDA appear alternately in different blocks.}
    \vspace{-4mm}
    \label{fig:arch}
\end{figure*}

\textbf{Substitutes of Self-attention.} As the core component of transformers, the self-attention module incurs the $O(N^2)$ computational/memory cost, where $N$ is the length of an embedding sequence.
Though such a cost may be acceptable for image classification, it is not the case for other tasks with much larger input images (\eg, object detection and segmentation).
To alleviate the cost, Swin~\citep{DBLP:journals/corr/abs-2103-14030} restricts the attention in a certain local region, giving up long-distance dependencies. PVT~\citep{DBLP:journals/corr/abs-2102-12122} and Twins~\citep{DBLP:journals/corr/abs-2104-13840} make adjacent embeddings share the same \textit{key/value} to reduce the cost.
Likewise, other vision transformers such as~\citep{DBLP:journals/corr/abs-2103-14899,DBLP:journals/corr/abs-2105-12723,DBLP:journals/corr/abs-2103-15808} also employ a divide-and-conquer method and approximate the vanilla self-attention module with a lower cost.

\textbf{Position Representations.} Transformer is combination-invariant. That is, shuffling input embeddings does not change the output of a transformer. Nevertheless, the position of embeddings also contains important information.
To make the respective model aware of this, many different position representations of embeddings~\citep{DBLP:conf/nips/VaswaniSPUJGKP17} are proposed. For example, ~\citet{DBLP:conf/iclr/DosovitskiyB0WZ21} directly add the embeddings with the vectors that contain absolute position information. In contrast, Relative Position Bias (RPB)~\citep{DBLP:conf/naacl/ShawUV18} resorts to position information indicating the relative distance of two embeddings.
Much recent work~\citep{DBLP:journals/corr/abs-2103-14030,DBLP:journals/corr/abs-2106-02689} shows that RPB performs better than other position representations. Motivated by this finding, our proposed position representation DPB also uses relative distance, but different from RPB that only handles fixed-sized images, our DPB applies to images with dynamic sizes.

\vspace{-4mm}
\section{CrossFormer}
\vspace{-3mm}

The overall architecture of CrossFormer is plotted in Figure~\ref{fig:arch}.
Following \citep{DBLP:journals/corr/abs-2102-12122,DBLP:journals/corr/abs-2103-14030,DBLP:journals/corr/abs-2106-05786}, CrossFormer also employs a pyramid structure, which naturally splits the transformer model into four stages.
Each stage consists of a cross-scale embedding layer (CEL, Section~\ref{sec:cel}) and several CrossFormer blocks (Section~\ref{sec:block}).
A CEL receives last stage's output (or an input image) as input and generates cross-scale embeddings.
In this process, CEL (except that in \textit{Stage-1}) reduces the number of embeddings to a quarter while doubles their dimensions for a pyramid structure.
Then, several CrossFormer blocks, each of which involves long short distance attention (LSDA) and dynamic position bias (DPB), are set up after CEL.
A specialized head (\eg, the classification head in Figure~\ref{fig:arch}) follows after the final stage accounting for a specific task.

\vspace{-1mm}
\subsection{Cross-scale Embedding Layer (CEL)} \label{sec:cel}
\begin{wrapfigure}[]{r}{0.48\linewidth}
    \vspace{-1.2mm}
    \includegraphics[width=1.0\linewidth]{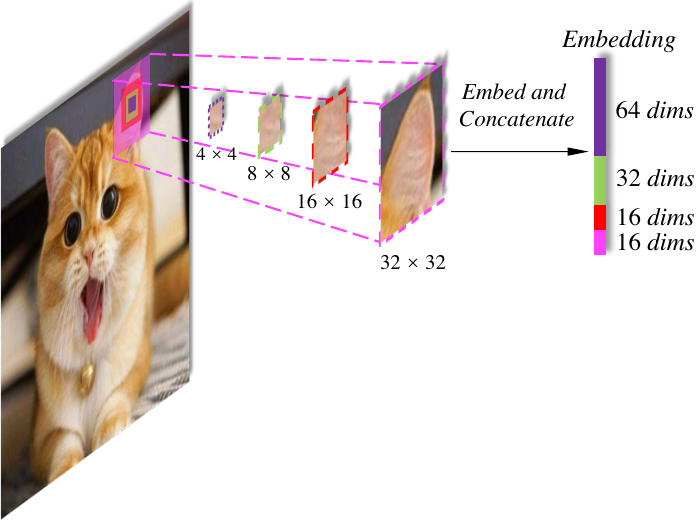}
    \hspace{-50mm}
    \renewcommand\arraystretch{1.25}
    \scalebox{0.8}{\setlength{\tabcolsep}{2.8mm}{\small
            \begin{tabular}[b]{cccc}
                \multicolumn{4}{c}{Embedding Layer} \\
                \toprule
                Type & Kernel & Stride & Dim \\
                \midrule
                Conv. & $4\times4$ & $4\times4$ & $\frac{D_t}{2}$\\
                Conv. & $8\times8$ & $4\times4$ & $\frac{D_t}{4}$ \\
                Conv. & $16\times16$ & $4\times4$ & $\frac{D_t}{8}$ \\
                Conv. & $32\times32$ & $4\times4$ & $\frac{D_t}{8}$ \\
                \bottomrule
    \end{tabular}}}
    \caption{Illustration of the CEL layer. The input image is sampled by four different kernels (\ie, $4 \times 4, 8 \times 8, 16 \times 16, 32 \times 32$) with same stride $4\times4$. Each embedding is constructed by projecting and concatenating the four patches. $D_t$ means the total dimension of the embedding.}
    \vspace{-3.5mm}
    \label{fig:embedding}
\end{wrapfigure}

\vspace{-1mm}
Cross-scale embedding layer (CEL) is leveraged to generate input embeddings for each stage.     
Figure~\ref{fig:embedding} takes the first CEL, which is ahead of \textit{Stage-1}, as an example.
It receives an image as input, then sampling patches using four kernels of different sizes.
The stride of four kernels is kept the same so that they generate the same number of embeddings\footnote{The image will be padded if necessary.}.
As we can observe in Figure~\ref{fig:embedding}, every four corresponding patches have the same center but different scales, and all these four patches will be projected and concatenated as one embedding.
In practice, the process of sampling and projecting can be fulfilled through four convolutional layers.

For a cross-scale embedding, one problem is how to set the projected dimension of each scale.
The computational budget of a convolutional layer is proportional to $K^2D^2$, where $K$ and $D$ represent kernel size and input/output dimension, respectively (assuming the input dimension equals to the output dimension).
Therefore, given the same dimension, a large kernel consumes more budget than a smaller one.
To control the total budget of the CEL, we use a lower dimension for large kernels while a higher dimension for small kernels.
Figure~\ref{fig:embedding} provides the specific allocation rule in its subtable, and a $128$ dimensional example is given.
Compared with allocating the dimension equally, our scheme saves much computational cost but does not explicitly affect the model's performance.
The cross-scale embedding layers in other stages work in a similar way. As shown in Figure~\ref{fig:arch}, CELs for \textit{Stage-2/3/4} use two different kernels ($2 \times 2$ and $4 \times 4$). Further, to form a pyramid structure, the strides of CELs for \textit{Stage-2/3/4} are set as $2 \times 2$ to reduce the number of embeddings to a quarter.

\vspace{-2mm}
\subsection{CrossFormer Block}  \label{sec:block}
\vspace{-1mm}

Each CrossFormer block consists of a long short distance attention module (\ie, LSDA, which involves a short distance attention (SDA) module or a long distance attention (LDA) module) and a multilayer perceptron (MLP).
As shown in Figure~\ref{fig:arch}b, SDA and LDA appear alternately in different blocks, and the dynamic position bias (DPB) module works in both SDA and LDA for obtaining embeddings' position representations.
Following the prior vision transformers, residual connections are used in each block.

\vspace{-2mm}
\subsubsection{Long Short Distance Attention (LSDA)}
\vspace{-1mm}

We split the self-attention module into two parts: short distance attention (SDA) and long distance attention (LDA).
For SDA, every $G \times G$ adjacent embeddings are grouped together. Figure~\ref{fig:LSDA}a gives an example where $G=3$.
For LDA with input of size $S \times S$, the embeddings are sampled with a fixed interval $I$. For example in Figure~\ref{fig:LSDA}b ($I=3$), all embeddings with a red border belong to a group, and those with a yellow border comprise another group. The group's height or width for LDA is computed as $G=S/I$ (\ie, $G=3$ in this example). After grouping embeddings, both SDA and LDA employ the vanilla self-attention within each group. As a result, the memory/computational cost of the self-attention module is reduced from $O(S^4)$ to $O(S^2G^2)$, and $G \ll S$ in most cases.

It is worth noting that the effectiveness of LDA also benefits from cross-scale embeddings. Specifically,
we draw all the patches comprising two embeddings in Figure~\ref{fig:LSDA}b.
As we can see,
the small-scale patches of two embeddings are non-adjacent, so it is difficult to judge their relationship without the help of the context.
In other words, it will be hard to build the dependency between these two embeddings if they are only constructed by small-scale patches (\ie, single-scale feature). On the contrary, adjacent large-scale patches provide sufficient context to link these two embeddings, which makes long-distance cross-scale attention easier and more meaningful.

We provide the pseudo-code of LSDA in the appendix (\ref{apd:pesudo}). Based on the vanilla multi-head self-attention, LSDA can be implemented with only ten lines of code. Further, only \textit{reshape} and \textit{permute} operations are used, so no extra computational cost is introduced.

\vspace{-2mm}
\subsubsection{Dynamic Position Bias (DPB)} \label{sec:dpb}
\vspace{-1mm}

\begin{figure*}[t]
    \centering
    \includegraphics[width=0.9\linewidth]{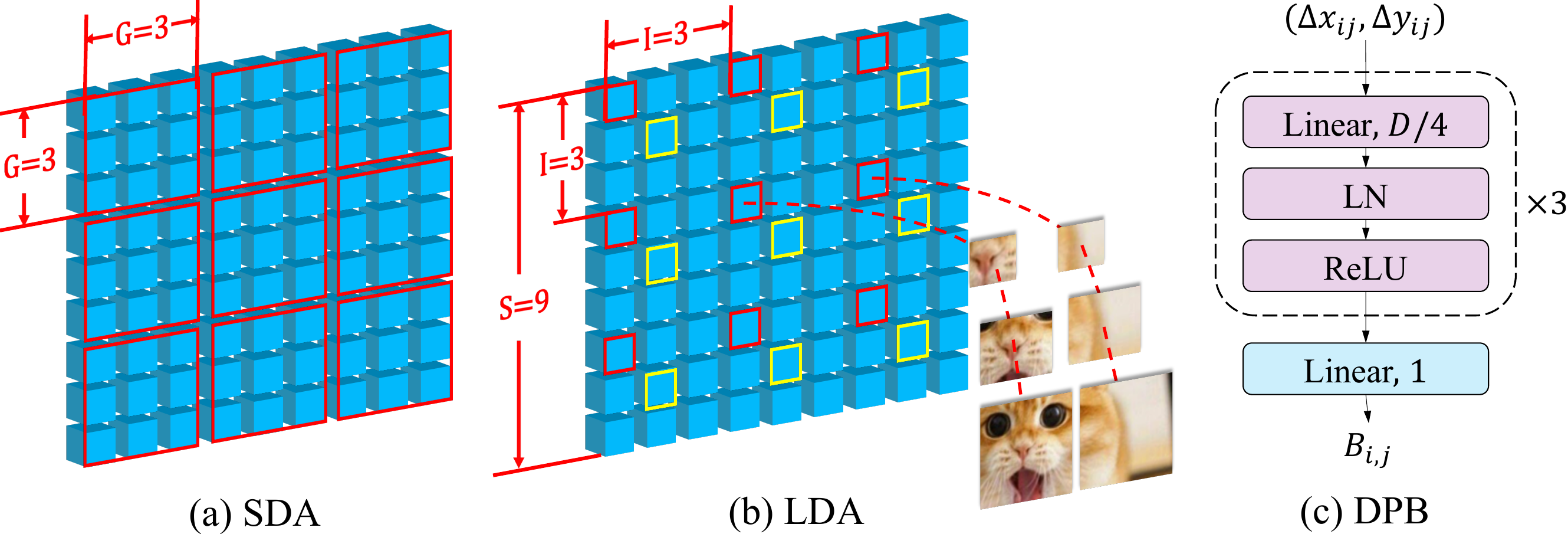}
    \caption{
        (a) Short distance attention (SDA). Embeddings (blue cubes) are grouped by red boxes.
        (b) Long distance attention (LDA). Embeddings with the same color borders belong to the same group. Large patches of embeddings in the same group are adjacent.
        (c) Dynamic position bias (DBP). The dimensions of intermediate layers are $D/4$, and the output is a scalar.}
    \label{fig:LSDA}
    \vspace{-3mm}
\end{figure*}

Relative position bias (RPB) indicates embeddings' relative position by adding a bias to their attention. Formally, the LSDA's attention map with RPB becomes:
\begin{equation}
    \label{equ:attn}
    \mathtt{Attention} = \mathtt{Softmax}(\mQ\mK^T/\sqrt d + \mB)\mV,
\end{equation}
where $\mQ, \mK, \mV \in \mathbb{R}^{G^2 \times D}$ represent \textit{query, key, value} in the self-attention module, respectively. $\sqrt{d}$ is a constant normalizer, and $\mB \in \mathbb{R}^{G^2 \times G^2}$ is the RPB matrix. In previous works~\citep{DBLP:journals/corr/abs-2103-14030}, $\mB_{i,j} = \hat{\mB}_{\Delta x_{ij}, \Delta y_{ij}}$, where $\hat{\mB}$ is a fixed-sized matrix, and $(\Delta x_{ij}, \Delta y_{ij})$ is the coordinate distance between the $i_{th}$ and $j_{th}$ embeddings. It is obvious that the image/group size is restricted in case that $(\Delta x_{ij}, \Delta y_{ij})$ exceeds the size of $\hat{B}$. In contrast, we propose an MLP-based module called DPB to generate the relative position bias dynamically, \ie,
\begin{equation}
    \label{equ:attn_bias}
    \mB_{i,j} = DPB(\Delta x_{ij}, \Delta y_{ij}).
\end{equation}
The structure of DPB is displayed in Figure~\ref{fig:LSDA}c. Its non-linear transformation consists of three fully-connected layers with layer normalization~\citep{DBLP:journals/corr/BaKH16} and ReLU~\citep{DBLP:conf/icml/NairH10}.
The input dimension of DPB is $2$, \ie, $(\Delta x_{ij}, \Delta y_{ij})$, and intermediate layers' dimension is set as $D/4$, where $D$ is the dimension of embeddings. The output $B_{ij}$ is a scalar, encoding the relative position feature between the $i_{th}$ and $j_{th}$ embeddings.
DPB is a trainable module optimized along with the whole transformer model. It can deal with any image/group size without worrying about the bound of $(\Delta x_{ij}, \Delta y_{ij})$.
In the appendix (\ref{apd:dpb}), we prove that DPB is equivalent to RPB if the image/group size is fixed. In this case, we can transform a trained DPB to RPB in the test phase. We also provide an efficient $O(G^2)$ implementation of DPB when image/group size is variable (the complexity is $O(G^4)$ in a normal case because $\mB \in \mathbb{R}^{G^2 \times G^2}$).

\begin{table}[]
    \centering
    \caption{Variants of CrossFormer for image classification. The example input size is $224 \times 224$. $S$ represents the feature maps' height (and width) of each stage. $D$ and $H$ mean embedding dimensions and the number of heads in the multi-head self-attention module, respectively. $G$ and $I$ are group size and interval for SDA and LDA, respectively.}
    \scalebox{0.7}{
        \setlength{\tabcolsep}{3mm}{
            \begin{tabular}{cc|c|cccc}
                \toprule
                & Output Size & Layer Name & CrossFormer-T & CrossFormer-S & CrossFormer-B & CrossFormer-L \\ \midrule
                \multirow{5}{*}{Stage-1} & \multirow{4}{*}{$ 56 \times 56 $} & Cross Embed. & \multicolumn{4}{c}{Kernel size: $4 \times 4$, $8 \times 8$, $16 \times 16$, $32 \times 32$, Stride=$4$} \\
                \cmidrule{3-7}
                & \multirow{4}{*}{$ (S_1 = 56) $} & \multirow{3}{*}{SDA/LDA} & \multirow{2}{*}{$\begin{bmatrix}D_1=64 \\H_1=2 \\G_1=7 \\I_1=8\end{bmatrix} \times 1$} & \multirow{2}{*}{$\begin{bmatrix}D_1=96 \\H_1=3 \\G_1=7 \\I_1=8\end{bmatrix} \times 2$} & \multirow{2}{*}{$\begin{bmatrix}D_1=96\\H_1=3\\G_1=7\\I_1=8, \end{bmatrix} \times 2$} & \multirow{2}{*}{$\begin{bmatrix}D_1=128 \\ H_1=4 \\ G_1=7 \\ I_1=8\end{bmatrix} \times 2$} \\
                &  & \multirow{3}{*}{MLP} &  &  &  \\
                &  &  & & & & \\
                &  &  & & & & \\
                \midrule
                \multirow{5}{*}{Stage-2} & \multirow{4}{*}{$ 28\times28 $} & Cross Embed. & \multicolumn{4}{c}{Kernel size: $2 \times 2$, $4 \times 4$, Stride=$2$} \\
                \cmidrule{3-7}
                & \multirow{4}{*}{$ (S_2 = 28) $} & \multirow{3}{*}{SDA/LDA} & \multirow{3}{*}{$\begin{bmatrix}D_2=128\\H_2=4 \\ G_2=7\\I_2= 4 \end{bmatrix} \times 1$} & \multirow{2}{*}{$\begin{bmatrix}D_2=192\\H_2=6 \\ G_2=7\\I_2= 4 \end{bmatrix} \times 2$} & \multirow{2}{*}{$\begin{bmatrix}D_2=192\\H_2=6 \\ G_2=7\\I_2=4 \end{bmatrix} \times 2$} & \multirow{2}{*}{$\begin{bmatrix}D_2=256\\H_2=8 \\ G_2=7\\I_2=4 \end{bmatrix} \times 2$} \\
                &  & \multirow{3}{*}{MLP} &  &  &  \\
                &  &  & & & & \\
                &  &  & & & & \\
                \midrule
                \multirow{5}{*}{Stage-3} & \multirow{4}{*}{$ 14\times14 $} & Cross Embed. & \multicolumn{4}{c}{Kernel size: $2 \times 2$, $4 \times 4$, Stride=$2$} \\
                \cmidrule{3-7}
                & \multirow{4}{*}{$ (S_3 = 14) $} & \multirow{3}{*}{SDA/LDA} & \multirow{3}{*}{$\begin{bmatrix}D_3=256\\H_3=8 \\ G_3=7\\I_3=2 \end{bmatrix} \times 8$} & \multirow{2}{*}{$\begin{bmatrix}D_3=384\\H_3=12 \\ G_3=7\\I_3=2 \end{bmatrix} \times 6$} & \multirow{2}{*}{$\begin{bmatrix}D_3=384\\H_3=12 \\ G_3=7\\I_3=2 \end{bmatrix} \times 18$} & \multirow{2}{*}{$\begin{bmatrix}D_3=512\\H_3=16 \\ G_3=7\\I_3=2 \end{bmatrix} \times 18$} \\
                &  & \multirow{3}{*}{MLP} &  &  &  \\
                &  &  & & & & \\
                &  &  & & & & \\
                \midrule
                \multirow{5}{*}{Stage-4} & \multirow{4}{*}{$ 7\times7 $} & Cross Embed. & \multicolumn{4}{c}{Kernel size: $2 \times 2$, $4 \times 4$, Stride=$2$} \\
                \cmidrule{3-7}
                & \multirow{4}{*}{$ (S_4 = 7) $} & \multirow{3}{*}{SDA/LDA} & \multirow{3}{*}{$\begin{bmatrix}D_4=512\\H_4=16 \\ G_4=7\\I_4=1 \end{bmatrix} \times 6$} & \multirow{2}{*}{$\begin{bmatrix}D_4=768\\H_4=24 \\ G_4=7\\I_4=1 \end{bmatrix} \times 2$} & \multirow{2}{*}{$\begin{bmatrix}D_4=768\\H_4=24 \\ G_4=7\\I_4=1 \end{bmatrix} \times 2$} & \multirow{2}{*}{$\begin{bmatrix}D_4=1024\\H_4=32 \\ G_4=7\\I_4=1 \end{bmatrix} \times 2$} \\
                &  & \multirow{3}{*}{MLP} &  &  &  \\
                &  &  & & & & \\
                &  &  & & & & \\
                \midrule
                \multirow{2}{*}{Head} & \multirow{2}{*}{$ 1 \times 1 $} & Avg Pooling & \multicolumn{4}{c}{Kernel size: $7 \times 7$} \\
                \cmidrule{3-7}
                &  & Linear & \multicolumn{4}{c}{Classes: $ 1000 $} \\
                \bottomrule
    \end{tabular}}}
    \label{tab:variants}
       \vspace{-4mm}
\end{table}

\vspace{-2mm}
\subsection{Variants of CrossFormer}
\vspace{-1mm}

Table~\ref{tab:variants} lists the detailed configurations of CrossFormer's four variants (-T, -S, -B, and -L for tiny, small, base, and large, respectively) for image classification.
To re-use the pre-trained weights, the models for other tasks (\eg, object detection) employ the same backbones as classification except that they may use different $G$ and $I$. Specifically, besides the configurations same to classification, we also test with $G_1=G_2=14, I_1 = 16$, and $I_2 = 8$ for the detection (and segmentation) models' first two stages to adapt to larger images. The specific configurations are described in the appendix (\ref{apd:variants}).
Notably, the group size or the interval (\ie, $G$ or $I$) does not affect the shape of weight tensors, so the backbones pre-trained on ImageNet can be readily fine-tuned on other tasks even though they use different $G$ or $I$.

\vspace{-2mm}
\section{Experiments}
\vspace{-1mm}

The experiments are carried out on four challenging tasks: image classification, object detection, instance segmentation, and semantic segmentation. To entail a fair comparison, we keep the same data augmentation and training settings as the other vision transformers as far as possible.
The competitors are all competitive vision transformers, including DeiT~\citep{DBLP:conf/icml/TouvronCDMSJ21}, PVT~\citep{DBLP:journals/corr/abs-2102-12122}, T2T-ViT~\citep{DBLP:journals/corr/abs-2101-11986}, TNT~\citep{DBLP:journals/corr/abs-2103-00112}, CViT~\citep{DBLP:journals/corr/abs-2103-14899}, Twins~\citep{DBLP:journals/corr/abs-2104-13840}, Swin~\citep{DBLP:journals/corr/abs-2103-14030}, NesT~\citep{DBLP:journals/corr/abs-2105-12723}, CvT~\citep{DBLP:journals/corr/abs-2103-15808},  ViL~\citep{DBLP:journals/corr/abs-2103-15358},
CAT~\citep{DBLP:journals/corr/abs-2106-05786}, ResT~\citep{DBLP:journals/corr/abs-2105-13677}, TransCNN~\citep{DBLP:journals/corr/abs-2106-03180}, Shuffle~\citep{DBLP:journals/corr/abs-2106-03650}, BoTNet~\citep{DBLP:journals/corr/abs-2101-11605}, and RegionViT~\citep{DBLP:journals/corr/abs-2106-02689}.

\vspace{-2mm}
\subsection{Image Classification}
\vspace{-1mm}
\textbf{Experimental Settings.} The experiments on image classification are done with the ImageNet~\citep{DBLP:journals/ijcv/RussakovskyDSKS15} dataset. The models are trained on $1.28$M training images and tested on $50$K validation images. The same training settings as the other vision transformers are adopted. In particular, we use an AdamW~\citep{DBLP:journals/corr/KingmaB14} optimizer training for 300 epochs with a cosine decay learning rate scheduler, and 20 epochs of linear warm-up are used. The batch size is 1,024 split on 8 V100 GPUs. An initial learning rate of 0.001 and a weight decay of 0.05 are used.
Besides, we use drop path rate of $0.1, 0.2, 0.3, 0.5$ for CrossFormer-T, CrossFormer-S, CrossFormer-B, CrossFormer-L, respectively. Further, Similar to Swin~\citep{DBLP:journals/corr/abs-2103-14030}, RandAugment~\citep{DBLP:conf/nips/CubukZS020}, Mixup~\citep{DBLP:conf/iclr/ZhangCDL18}, Cutmix~\citep{DBLP:conf/iccv/YunHCOYC19}, random erasing~\citep{DBLP:conf/aaai/Zhong0KL020}, and stochastic depth~\citep{DBLP:conf/eccv/HuangSLSW16} are used for data augmentation.

\textbf{Results.} The results are shown in Table \ref{tab:classification}. As we can see, CrossFormer achieves the highest accuracy with parameters and FLOPs comparable to other state-of-the-art vision transformer structures. In specific, compared against strong baselines DeiT, PVT, and Swin, our CrossFormer outperforms them at least absolute $1.2\%$ in accuracy on small models.
Further, though RegionViT achieves the same accuracy ($82.5\%$) as ours on a small model, it is $0.7\%$ ($84.0\%$ vs. $83.3\%$) absolutely lower than ours on the large model.

\begin{table}[]
    \centering
    \caption{Results on the ImageNet validation set. The input size is $224 \times 224$ for most models, while is $384 \times 384$ for the model with a $^\dagger$. Results of other architectures are drawn from original papers.}
    \scalebox{0.8}{
        \setlength{\tabcolsep}{3mm}{
            \begin{subtable}[h]{0.46\textwidth}
                \begin{tabular}{l|rrrr}
                    \toprule
                    Architectures& \#Params & FLOPs & Acc. \\
                    \midrule
                    PVT-S & 24.5M & 3.8G & 79.8\% \\
                    RegionViT-T & 13.8M & 2.4G & 80.4\% \\
                    Twins-SVT-S & 24.0M & 2.8G & 81.3\%  \\% & 56.0M & 8.3G & 83.1\%  & 99.2M & 14.8G & 83.3\% \\
                    \textbf{CrossFormer-T} & 27.8M & 2.9G & \textbf{81.5\%} \\
                    \midrule
                    DeiT-S & 22.1M & 4.6G & 79.8\% \\% & 86.0M & 17.5G & 81.8\% & 86.0M & 55.4G & $^\dagger$83.1\% \\
                    T2T-ViT & 21.5M & 5.2G & 80.7\% \\
                    CViT-S & 26.7M & 5.6G & 81.0\%  \\% & 43.3M & 9.0G & 82.5\%  & - & - & - \\
                    PVT-M & 44.2M & 6.7G & 81.2\%  \\% & 61.4M & 9.8G & 81.7\%  & - & - & - \\
                    TNT-S & 23.8M & 5.2G & 81.3\% \\
                    Swin-T & 29.0M & 4.5G & 81.3\%  \\% & 50.0M & 8.7G & 83.0\%  & 88.0M & 15.4G & 83.3\% \\
                    NesT-T & 17.0M & 5.8G & 81.5\%  \\% & 38.0M & 10.4G & 83.3\%  & 68.0M & 17.9G & 83.8\% \\
                    CvT-13& 20.0M & 4.5G & 81.6\%  \\% & 32.0M & 7.1G & 82.5\%  & - & - & - \\
                    ResT & 30.2M & 4.3G & 81.6\% \\
                    CAT-S & 37.0M & 5.9G & 81.8\%  \\% & 52.0M & 8.9G & 82.8\%  & - & - & - \\
                    ViL-S & 24.6M & 4.9G & 81.8\%  \\% & 55.7 & 13.4 & 83.2  & - & - & - \\
                    RegionViT-S & 30.6M & 5.3G & \textbf{82.5\%} \\% & 41.2M & 7.4G & 83.1\%  & 72.0M & 13.3G & 83.3\% \\
                    \textbf{CrossFormer-S} & 30.7M & 4.9G & \textbf{82.5\%}  \\% & 52.0M & 9.2G & \textbf{83.4\%}  & 92.0M & 16.1G & \textbf{84.0\%} \\
                    \bottomrule
                \end{tabular}
        \end{subtable}}
        \hspace{20mm}
        \setlength{\tabcolsep}{3mm}{
            \begin{subtable}[h]{0.57\textwidth}
                \begin{tabular}{l|rrrr}
                    \toprule
                    Architectures& \#Params & FLOPs & Acc. \\
                    \midrule
                    BoTNet-S1-59 & 33.5M & 7.3G & 81.7\% \\
                    PVT-L & 61.4M & 9.8G & 81.7\%  \\ % & - & - & - \\
                    CvT-21 & 32.0M & 7.1G & 82.5\%  \\ % & - & - & - \\
                    CAT-B & 52.0M & 8.9G & 82.8\%  \\ % & - & - & - \\
                    Swin-S & 50.0M & 8.7G & 83.0\%  \\ % & 88.0M & 15.4G & 83.3\% \\
                    RegionViT-M & 41.2M & 7.4G & 83.1\%  \\ % & 72.0M & 13.3G & 83.3\% \\
                    Twins-SVT-B  & 56.0M & 8.3G & 83.1\% \\ %   & 99.2M & 14.8G & 83.3\% \\
                    NesT-S & 38.0M & 10.4G & 83.3\% \\ % & 68.0M & 17.9G & 83.8\% \\
                    \textbf{CrossFormer-B} & 52.0M & 9.2G & \textbf{83.4\%}  \\ % & 92.0M & 16.1G & \textbf{84.0\%} \\
                    \midrule
                    DeiT-B & 86.0M & 17.5G & 81.8\% \\ % & 86.0M & 55.4G & $^\dagger$83.1\% \\
                    DeiT-B$^\dagger$ & 86.0M & 55.4G & 83.1\% \\
                    ViL-B & 55.7M & 13.4G & 83.2\% \\
                    RegionViT-B & 72.0M & 13.3G & 83.3\% \\
                    Twins-SVT-L & 99.2M & 14.8G & 83.3\% \\
                    Swin-B & 88.0M & 15.4G & 83.3\% \\
                    NesT-B & 68.0M & 17.9G & 83.8\% \\
                    \textbf{CrossFormer-L} & 92.0M & 16.1G & \textbf{84.0\%} \\
                    \bottomrule
                \end{tabular}
    \end{subtable}}}
    \vspace{-4mm}
    \label{tab:classification}
\end{table}

\vspace{-2mm}
\subsection{Object Detection and Instance Segmentation}
\vspace{-1mm}

\textbf{Experimental Settings.} The experiments on object detection and instance segmentation are both done on the COCO 2017 dataset~\citep{DBLP:conf/eccv/LinMBHPRDZ14}, which contains $118$K training and $5$K val images.
We use MMDetection-based~\citep{mmdetection} RetinaNet~\citep{DBLP:journals/pami/LinGGHD20} and Mask R-CNN~\citep{DBLP:conf/iccv/HeGDG17} as the object detection and instance segmentation head, respectively.
For both tasks, the backbones are initialized with the weights pre-trained on ImageNet. Then the whole models are trained with batch size $16$ on $8$ V100 GPUs, and an AdamW optimizer with an initial learning rate of $1 \times 10^{-4}$ is used. Following previous works, we adopt $1\times$ training schedule (\ie, the models are trained for $12$ epochs) when taking RetinaNets as detectors, and images are resized to 800 pixels for the short side. While for Mask R-CNN, both $1\times$ and $3 \times$ training schedules are used. It is noted that multi-scale training~\citep{DBLP:conf/eccv/CarionMSUKZ20} is also employed when taking $3 \times$ training schedules.
\begin{table}[]
    \centering
    \caption{Object detection results on COCO 2017 \textit{val} set with RetinaNets as detectors. Results for Swin are drawn from Twins as Swin does not report results on RetinaNet. Results in \textcolor{blue}{blue} fonts are the \textcolor{blue}{second-placed} ones. CrossFormers with $^\ddagger$ use different group sizes from classification models. (More details are put in the appendix (\ref{apd:variants}))}
    \scalebox{0.8}{\setlength{\tabcolsep}{2.8mm}{
            \begin{tabular}{c|l|rr|lll|lll}
                \toprule
                Method & Backbone & \#Params & FLOPs & AP$^b$ & AP$^b_{50}$ & AP$^b_{75}$ & AP$^b_{S}$ & AP$^b_{M}$ & AP$^b_{L}$  \\
                \midrule
                \multirow{16}{*}{RetinaNet} & ResNet-50 & 37.7M & 234.0G & 36.3 & 55.3 & 38.6 & 19.3 & 40.0 & 48.8  \\
                \multirow{16}{*}{$1\times$ schedule} & CAT-B & 62.0M & 337.0G & 41.4 & 62.9& 43.8 & 24.9 & 44.6 & 55.2 \\
                & Swin-T & 38.5M & 245.0G  & 41.5 & 62.1 & 44.2 & 25.1 & 44.9 & 55.5 \\
                & PVT-M & 53.9M & $-$ & 41.9 & 63.1 & 44.3 & 25.0 & 44.9 & 57.6 \\
                & ViL-M & 50.8M & 338.9G & 42.9 & 64.0 & 45.4 & 27.0 & 46.1 & 57.2 \\
                & RegionViT-B & 83.4M & 308.9G & 43.3 & 65.2 & 46.4 & 29.2 & 46.4 & 57.0 \\
                & \textcolor{blue}{TransCNN-B} & \textcolor{blue}{36.5M} & $-$ & \textcolor{blue}{43.4} & 64.2 & 46.5 & 27.0 & 47.4 & 56.7 \\
                & \textbf{CrossFormer-S} & 40.8M & 282.0G & \textbf{44.4}\,\textcolor{blue}{(+1.0)} & 65.8 & 47.4 & 28.2 & 48.4 & 59.4 \\
                & \textbf{CrossFormer-S$^\ddagger$} & 40.8M & 272.1G & \textbf{44.2}\,\textcolor{blue}{(+0.8)} & 65.7 & 47.2 & 28.0 & 48.0 & 59.1 \\
                \cmidrule{2-10}
                & ResNet101 & 56.7M & 315.0G & 38.5 & 57.8 & 41.2 & 21.4 & 42.6 & 51.1 \\
                & PVT-L & 71.1M & 345.0G & 42.6 & 63.7 & 45.4 & 25.8 & 46.0 & 58.4 \\
                & Twins-SVT-B & 67.0M & 322.0G & 44.4 & 66.7 & 48.1 & 28.5 & 48.9 & 60.6 \\
                & RegionViT-B+ & 84.5M & 328.2G & 44.6 & 66.4 & 47.6 & 29.6 & 47.6 & 59.0 \\
                & Swin-B & 98.4M & 477.0G & 44.7 & 65.9 & 49.2 & $-$ & $-$ & $-$ \\
                & \textcolor{blue}{Twins-SVT-L} & \textcolor{blue}{110.9M} & \textcolor{blue}{455.0G} & \textcolor{blue}{44.8} & 66.1 & 48.1 & 28.4 & 48.3 & 60.1 \\
                & \textbf{CrossFormer-B} & 62.1M & 389.0G & \textbf{46.2}\,\textcolor{blue}{(+1.4)} & 67.8 & 49.5 & 30.1 & 49.9 & 61.8 \\
                & \textbf{CrossFormer-B$^\ddagger$} & 62.1M & 379.1G & \textbf{46.1}\,\textcolor{blue}{(+1.3)} & 67.7 & 49.0 & 29.5 & 49.9 & 61.5 \\
                \bottomrule
    \end{tabular}}}
    \label{tab:detection}
        \vspace{-6mm}
\end{table}

\textbf{Results.} The results on RetinaNet and Mask R-CNN are shown in Table~\ref{tab:detection} and Table~\ref{tab:detection-2}, respectively. As we can see, the second-placed architecture changes along with the experiment, that is, these architectures may perform well on one task but poorly on another task. In contrast, our CrossFormer outperforms all the others on both tasks (detection and segmentation) with both model sizes (small and base). Further, CrossFormer's performance gain over the other architectures gets sharper when enlarging the model, indicating that CrossFormer enjoys greater potentials.

\begin{table}[]
    \centering
    \caption{Object detection and instance segmentation results on COCO \textit{val} 2017 with Mask R-CNNs as detectors. AP$^b$ and AP$^m$ are box average precision and mask average precision, respectively.}
    \scalebox{0.8}{
        \begin{tabular}{c|l|rr|lll|lll}
            \toprule
            Method & Backbone & \#Params & FLOPs & AP$^b$ & AP$^b_{50}$ & AP$^b_{75}$ & AP$^m$ & AP$^m_{50}$ & AP$^m_{75}$  \\
            \midrule
            \multirow{17}{*}{Mask R-CNN} & PVT-M & 63.9M & $-$ & 42.0 & 64.4 & 45.6 & 39.0 & 61.6 & 42.0 \\
            \multirow{17}{*}{$1\times$ schedule} & Swin-T & 47.8M & 264.0G & 42.2 & 64.6 & 46.2 & 39.1 & 61.6 & 42.0  \\
            & Twins-PCPVT-S & 44.3M & 245.0G & 42.9 & 65.8 & 47.1 & 40.0 & 62.7 & 42.9 \\
            & TransCNN-B & 46.4M & $-$ & 44.0 & 66.4 & 48.5 & 40.2 & 63.3 & 43.2 \\
            & ViL-M & 60.1M & 261.1G & 43.3 & 65.9 & 47.0 & 39.7 & 62.8 & 42.0 \\
            & RegionViT-B & 92.2M & 287.9G & 43.5 & 66.7 & 47.4 & 40.1 & 63.4 & 43.0 \\
            & \textcolor{blue}{RegionViT-B+} & \textcolor{blue}{93.2M} & \textcolor{blue}{307.1G}   & \textcolor{blue}{44.5} & 67.6 & 48.7 & \textcolor{blue}{41.0} & 64.4 & 43.9\\
            & \textbf{CrossFormer-S} & 50.2M & 301.0G & \textbf{45.4}\,\textcolor{blue}{(+0.9)} & 68.0 & 49.7 & \textbf{41.4}\,\textcolor{blue}{(+0.4)} & 64.8 & 44.6 \\
            & \textbf{CrossFormer-S$^\ddagger$} & 50.2M & 291.1G & \textbf{45.0}\,\textcolor{blue}{(+0.5)} & 67.9 & 49.1 & \textbf{41.2}\,\textcolor{blue}{(+0.2)} & 64.6 & 44.3 \\
            \cmidrule{2-10}
            & CAT-B & 71.0M & 356.0G & 41.8 & 65.4 & 45.2 & 38.7 & 62.3 & 41.4 \\
            & PVT-L & 81.0M & 364.0G & 42.9 & 65.0 & 46.6 & 39.5 & 61.9 & 42.5 \\
            & Twins-SVT-B & 76.3M & 340.0G & 45.1 & 67.0 & 49.4 & 41.1 & 64.1 & 44.4 \\
            & ViL-B & 76.1M & 365.1G & 45.1 & 67.2 & 49.3 & 41.0 & 64.3 & 44.2 \\
            & Twins-SVT-L & 119.7M & 474.0G & 45.2 & 67.5 & 49.4 & 41.2 & 64.5 & 44.5 \\
            & Swin-S & 69.1M & 354.0G & 44.8 & 66.6 & 48.9 & 40.9 & 63.4 & 44.2 \\
            & \textcolor{blue}{Swin-B} & \textcolor{blue}{107.2M} & \textcolor{blue}{496.0G} & \textcolor{blue}{45.5} & $-$ & $-$ & \textcolor{blue}{41.3} & $-$ & $-$ \\
            & \textbf{CrossFormer-B} & 71.5M & 407.9G & \textbf{47.2}\,\textcolor{blue}{(+1.7)} & 69.9 & 51.8 & \textbf{42.7}\,\textcolor{blue}{(+1.4)} & 66.6 & 46.2 \\
            & \textbf{CrossFormer-B$^\ddagger$} & 71.5M & 398.1G & \textbf{47.1}\,\textcolor{blue}{(+1.6)} & 69.9 & 52.0 & \textbf{42.7}\,\textcolor{blue}{(+1.4)} & 66.5 & 46.1 \\
            
            \midrule
            
            \multirow{10}{*}{Mask R-CNN} & PVT-M & 63.9M & $-$ & 44.2 & 66.0 & 48.2 & 45.0 & 63.1 & 43.5 \\
            \multirow{10}{*}{$3\times$ schedule} & ViL-M & 60.1M & 261.1G & 44.6 & 66.3 & 48.5 & 40.7 & 63.8 & 43.7 \\
            & Swin-T & 47.8M & 264.0G & 46.0 & 68.2 & 50.2 & 41.6 & 65.1 & 44.8 \\
            & \textcolor{blue}{Shuffle-T} & \textcolor{blue}{48.0M} & \textcolor{blue}{268.0G} & \textcolor{blue}{46.8} & 68.9 & 51.5 & \textcolor{blue}{42.3} & 66.0 & 45.6 \\
            & \textbf{CrossFormer-S$^\ddagger$}& 50.2M & 291.1G & \textbf{48.7}\,\textcolor{blue}{(+1.9)} & 70.7 & 53.7 & \textbf{43.9}\,\textcolor{blue}{(+1.6)} & 67.9 & 47.3 \\
            \cmidrule{2-10}
            & PVT-L & 81.0M & 364.0G & 44.5 & 66.0 & 48.3 & 40.7 & 63.4 & 43.7  \\
            & ViL-B & 76.1M & 365.1G & 45.7 & 67.2 & 49.9 & 41.3 & 64.4 & 44.5 \\
            & Shuffle-S & 69.0M & 359.0G & 48.4 & 70.1 & 53.5 & 43.3 & 67.3 & 46.7 \\
            & \textcolor{blue}{Swin-S} & \textcolor{blue}{69.1M} & \textcolor{blue}{354.0G} & \textcolor{blue}{48.5} & 70.2 & 53.5 & \textcolor{blue}{43.3} & 67.3 & 46.6 \\
            & \textbf{CrossFormer-B$^\ddagger$} & 71.5M & 398.1G & \textbf{49.8}\,\textcolor{blue}{(+1.3)} & 71.6 & 54.9 & \textbf{44.5}\,\textcolor{blue}{(+1.2)} & 68.8 & 47.9 \\
            
            \bottomrule
    \end{tabular}}
    \vspace{-2mm}
    \label{tab:detection-2}
\end{table}

\vspace{-2mm}
\subsection{Semantic Segmentation}
\vspace{-1mm}

\textbf{Experimental Settings.} ADE20K~\citep{DBLP:conf/cvpr/ZhouZPFB017} is used as the benchmark for semantic segmentation. It covers a broad range of $150$ semantic categories, including $20$K images for training and $2$K for validation. Similar to models for detection, we initialize the backbones with weights pre-trained on ImageNet, and MMSegmentation-based~\citep{mmseg2020} semantic FPN and UPerNet~\citep{DBLP:conf/eccv/XiaoLZJS18} are taken as the segmentation head. For FPN~\citep{DBLP:conf/cvpr/KirillovGHD19}, we use an AdamW optimizer with learning rate and weight deacy of $1\times10^{-4}$. Models are trained for $80$K iterations with batch size $16$. For UPerNet, an AdamW optimizer with an initial learning rate of $6\times10^{-5}$ and a weight decay of $0.01$ is used, and models are trained for $160$K iterations. %  with batch size $16$.

\begin{table}[]
    \caption{Semantic segmentation results on the ADE20K validation set. ``MS IOU'' means testing with variable input size.}
    \begin{subtable}[t]{0.4\textwidth}
        \centering
        \scalebox{0.78}{
            \begin{tabular}{l|rrl}
                \toprule
                \multicolumn{4}{c}{Semantic FPN ($80$K iterations)} \\
                Backbone & \#Params & FLOPs & IOU \\
                \midrule
                PVT-M & 48.0M & 219.0G & 41.6  \\
                Twins-SVT-B & 60.4M & 261.0G & 45.0 \\
                \textcolor{blue}{Swin-S} & \textcolor{blue}{53.2M} & \textcolor{blue}{274.0G} & \textcolor{blue}{45.2}  \\
                \textbf{CrossFormer-S} & 34.3M & 220.7G & \textbf{46.0}\,\textcolor{blue}{(+0.8)}  \\
                \textbf{CrossFormer-S$^\ddagger$} & 34.3M & 209.8G & \textbf{46.4}\,\textcolor{blue}{(+1.2)}  \\
                \midrule
                PVT-L & 65.1M & 283.0G & 42.1 \\
                \textcolor{blue}{CAT-B} & \textcolor{blue}{55.0M} & \textcolor{blue}{276.0G} & \textcolor{blue}{43.6} \\
                \textbf{CrossFormer-B} & 55.6M & 331.0G & \textbf{47.7}\,\textcolor{blue}{(+4.1)}   \\
                \textbf{CrossFormer-B$^\ddagger$} & 55.6M & 320.1G & \textbf{48.0}\,\textcolor{blue}{(+4.4)}   \\
                \midrule
                \textcolor{blue}{Twins-SVT-L} & \textcolor{blue}{103.7M} & \textcolor{blue}{397.0G} & \textcolor{blue}{45.8} \\
                \textbf{CrossFormer-L} & 95.4M & 497.0G & \textbf{48.7}\,\textcolor{blue}{(+2.9)}   \\
                \textbf{CrossFormer-L$^\ddagger$} & 95.4M & 482.7G & \textbf{49.1}\,\textcolor{blue}{(+3.3)}   \\
                \bottomrule
        \end{tabular}}
    \end{subtable}
    \hfill
    \begin{subtable}[t]{0.53\textwidth}
        \centering
        \scalebox{0.78}{
            \begin{tabular}{l|rrll}
                \toprule
                \multicolumn{5}{c}{UPerNet ($160$K iterations)} \\
                Backbone & \#Params & FLOPs & IOU & MS IOU\\
                \midrule
                Swin-T & 60.0M & 945.0G & 44.5 & 45.8 \\
                \textcolor{blue}{Shuffle-T} & \textcolor{blue}{60.0M} & \textcolor{blue}{949.0G} & \textcolor{blue}{46.6} & \textcolor{blue}{47.6} \\
                \textbf{CrossFormer-S} & 62.3M & 979.5G & \textbf{47.6}\,\textcolor{blue}{(+1.0)} & \textbf{48.4} \\
                \textbf{CrossFormer-S$^\ddagger$} & 62.3M & 968.5G & \textbf{47.4}\,\textcolor{blue}{(+0.8)} & \textbf{48.2} \\
                \midrule
                Swin-S & 81.0M & 1038.0G & 47.6 & 49.5 \\
                \textcolor{blue}{Shuffle-S} & \textcolor{blue}{81.0M} & \textcolor{blue}{1044.0G} & \textcolor{blue}{48.4} & 49.6 \\
                \textbf{CrossFormer-B} & 83.6M & 1089.7G  & \textbf{49.7}\,\textcolor{blue}{(+1.3)} & \textbf{50.6} \\
                \textbf{CrossFormer-B$^\ddagger$} & 83.6M & 1078.8G  & \textbf{49.2}\,\textcolor{blue}{(+0.8)} & \textbf{50.1} \\
                \midrule
                Swin-B & 121.0M & 1088.0G & 48.1 & 49.7 \\
                \textcolor{blue}{Shuffle-B} & \textcolor{blue}{121.0M} & \textcolor{blue}{1096.0G} & \textcolor{blue}{49.0} & $-$ \\
                \textbf{CrossFormer-L} & 125.5M & 1257.8G & \textbf{50.4}\,\textcolor{blue}{(+1.4)} & \textbf{51.4} \\
                \textbf{CrossFormer-L$^\ddagger$} & 125.5M & 1243.5G & \textbf{50.5}\,\textcolor{blue}{(+1.5)} & \textbf{51.4} \\
                \bottomrule
        \end{tabular}}
    \end{subtable}
    \label{tab:segmentation}
    \vspace{-2mm}
\end{table}

\textbf{Results.} All results are shown in Table~\ref{tab:segmentation}. Similar to object detection, CrossFormer exhibits a greater performance gain over the others when enlarging the model. For example, CrossFormer-T achieves $1.4\%$ absolutely higher on IOU than Twins-SVT-B, but CrossFormer-B achieves $3.1\%$ absolutely higher on IOU than Twins-SVT-L. Totally, CrossFormer shows a more significant advantage over the others on dense prediction tasks (\eg, detection and segmentation) than on classification, implying  that cross-scale interactions in the attention module are more important for dense prediction tasks than for classification.

\vspace{-2mm}
\subsection{Ablation Studies}
\vspace{-1mm}

\textbf{Cross-scale Embeddings vs. Single-scale Embeddings.} We conduct the experiments by replacing cross-scale embedding layers with single-scale ones.
As we can see in Table~\ref{tab:apd-classification}, when using single-scale embeddings, the $8 \times 8$ kernel in \textit{Stage-1} brings $0.4\%$ ($81.9\%$ vs. $81.5\%$) absolute improvement compared with the $4 \times 4$ kernel. It tells us that overlapping receptive fields help improve the model's performance. Besides, all models with cross-scale embeddings perform better than those with single-scale embeddings. In particular, our CrossFormer achieves $1\%$ ($82.5\%$ vs. $81.5\%$) absolute performance gain compared with using single-scale embeddings for all stages. For cross-scale embeddings, we also try several different combinations of kernel sizes, and they all show similar performance ($82.3\% \sim 82.5\%$). In summary, cross-scale embeddings can bring a large performance gain, yet the model is relatively robust to different choices of kernel size.

\textbf{LSDA vs. Other Self-attentions.} Two self-attention modules used in PVT and Swin are compared. Specifically, PVT sacrifices the small-scale features when computing the self-attention, while Swin restricts the self-attention in a local region, giving up the long-distance attention. As we can observe in Table~\ref{tab:ablation-1}, compared against the PVT-like and Swin-like self-attention mechanisms, our CrossFormer outperforms them at least absolute $0.6\%$ accuracy ($82.5\%$ vs. $81.9\%$). The results show that performing the self-attention in a long-short distance manner is most conducive to improving the model's performance.

\textbf{DPB vs. Other Position Representations.} We compare the parameters, FLOPs, throughputs, and accuracies of the models among absolute position embedding (APE), relative position bias (RPB), and DPB. The results are shown in Table~\ref{tab:ablation-3}. DPB-residual means DPB with residual connections. Both DPB and RPB outperform APE for absolute $0.4\%$ accuracy, which indicates that relative position representations are more beneficial than absolute ones.
Further, DPB achieves the same accuracy ($82.5\%$) as RPB with an ignorable extra cost; however, as we described in Section~\ref{sec:dpb}, it is more flexible than RPB and applies to variable image size or group size.
The results also show that residual connection in DPB does not help improve or even degrades the model's performance.

\begin{table}[t]
    \centering
    \caption{Results on the ImageNet validation set. The baseline model is CrossFormer-S (82.5\%). We test with different kernel sizes of CELs.}
    \scalebox{0.83}{
        \setlength{\tabcolsep}{2mm}{
            \begin{tabular}{c|c|c|c|rrr}
                \toprule
                \multicolumn{4}{c|}{CEL's Kernel Size} & \multirow{2}{*}{\#Params} & \multirow{2}{*}{FLOPs} & \multirow{2}{*}{Acc.} \\
                \textit{Stage-1} & \textit{Stage-2} & \textit{Stage-3} & \textit{Stage-4} & & & \\
                \midrule
                $4 \times 4$ & $2 \times 2$ & $2 \times 2$ & $2 \times 2$ & 28.3M & 4.5G & 81.5\% \\
                $8 \times 8$ & $2 \times 2$ & $2 \times 2$ & $2 \times 2$ & 28.3M & 4.5G & 81.9\% \\
                $4 \times 4, 8 \times 8$ & $2 \times 2, 4 \times 4$ & $2 \times 2, 4 \times 4$ & $2 \times 2, 4 \times 4$ & 30.6M & 4.8G & 82.3\% \\
                $4 \times 4, 8 \times 8, 16 \times 16, 32 \times 32$ & $2 \times 2, 4 \times 4$ & $2 \times 2, 4 \times 4$ & $2 \times 2, 4 \times 4$ & 30.7M & 4.9G & \textbf{82.5\%} \\
                $4 \times 4, 8 \times 8, 16 \times 16, 32 \times 32$ & $2 \times 2, 4 \times 4, 8 \times 8$ & $2 \times 2, 4 \times 4$ & $2 \times 2$ & 29.4M & 5.0G & 82.4\% \\
                \bottomrule
        \end{tabular}}
    }
    \label{tab:apd-classification}
\end{table}

\begin{table}[t]
    \centering
    \caption{Experimental results of ablation studies.}
    \vspace{-2mm}
    \setlength{\tabcolsep}{2.4mm}{
        \begin{subtable}[h]{0.45\textwidth}
            \centering
            \caption{Ablation studies on cross-scale embeddings (CEL) and long short distance attention (LSDA). The base model is CrossFormer-S (82.5\%).}
            \scalebox{0.78}{
                \begin{tabular}{ccc|c|c}
                    \toprule
                    PVT-like & Swin-like & LSDA & CEL & Acc. \\
                    \midrule
                    \checkmark &  & & \checkmark & \textbf{81.3\%}  \\
                    & \checkmark & & \checkmark & \textbf{81.9\%}  \\
                    &  & \checkmark & \checkmark & \textbf{82.5\%}  \\
                    &  & \checkmark & & \textbf{81.5\%}  \\
                    \bottomrule
            \end{tabular}}
            \label{tab:ablation-1}
        \end{subtable}
        \hspace{1mm}
        \begin{subtable}[h]{0.5\textwidth}
            \caption{Comparisons between different position representations. The base model is CrossFormer-S. Throughput is tested on $1\times$ V100 GPU.}
            \scalebox{0.78}{
                \begin{tabular}{lccc}
                    \toprule
                    Method& \#Params/FLOPs & Throughput & Acc. \\
                    \midrule
                    APE & 30.9342M/4.9061G & 686 imgs/sec & 82.1\% \\
                    RPB & 30.6159M/4.9062G & 684 imgs/sec & \textbf{82.5\%} \\
                    DPB & 30.6573M/4.9098G & 672 imgs/sec & \textbf{82.5\%} \\
                    DPB-residual & 30.6573M/4.9098G & 672 imgs/sec & 82.4\% \\
                    \bottomrule
            \end{tabular}}
            \label{tab:ablation-3}
        \end{subtable}
    }
    \label{tab:ablation}
    \vspace{-2mm}
\end{table}

\vspace{-3mm}
\section{Conclusions}
\vspace{-2mm}

We proposed a novel transformer-based vision architecture, namely CrossFormer. Its core ingredients are Cross-scale Embedding Layer (CEL) and Long Short Distance Attention (LSDA), thereby yielding the cross-attention module. We further proposed a dynamic position bias, making the relative position bias apply to any input size. Extensive experiments show that CrossFormer achieves superior performance over other state-of-the-art vision transformers on several representative vision tasks. Particularly, CrossFormer is demonstrated to gain great improvements on object detection and segmentation, which indicates that CEL and LSDA are together essential for dense prediction tasks.

\bibliography{CrossFormer}
\bibliographystyle{iclr2022_conference}
\newpage

\appendix
\section{CrossFormer}

\subsection{Pseudo code of LSDA} \label{apd:pesudo}
The pseudo code for LSDA is shown in Algorithm~\ref{alg:lsda}. As we can see, based on the vanilla self-attention module, both SDA and LDA are implemented with only ten lines of code, and only \textit{reshape} and \textit{permute} operations are used.
\begin{algorithm}[]
\caption{{LSDA code (PyTorch-like)}}
\label{alg:lsda}
\definecolor{codeblue}{rgb}{0.25,0.5,0.25}
\lstset{
    backgroundcolor=\color{white},
    basicstyle=\fontsize{7.2pt}{7.2pt}\ttfamily\selectfont,
    columns=fullflexible,
    breaklines=true,
    captionpos=b,
    commentstyle=\fontsize{7.2pt}{7.2pt}\color{codeblue},
    keywordstyle=\fontsize{7.2pt}{7.2pt},
}
\begin{lstlisting}[language=python]
# H: height, W: width, G: group size of SDA/LDA
# x: input tensor (H, W, D)
class LSDA():  
    def forward(x, type):
        ## group the embeddings
        if type == "SDA":
            x = x.reshaspe(H // G, G, W // G, G, D).permute(0, 2, 1, 3, 4)
        elif type == "LDA":
            x = x.reshaspe(G, H // G, G, W // G, D).permute(1, 3, 0, 2, 4)
        x = x.reshape(H * W // (G ** 2), G ** 2, D)
    
        ## the vanilla self-attention module
        x = Attention(x)
    
        ## un-group the embeddings
        x = x.reshaspe(H // G, W // G, G, G, D)
        if type == "SDA":
            x = x.permute(0, 2, 1, 3, 4).reshaspe(H, W, D)
        elif type == "LDA":
            x = x.permute(2, 0, 3, 1, 4).reshaspe(H, W, D)
        return x
\end{lstlisting}
\end{algorithm}
    
\subsection{Dynamic Position Bias (DPB)} \label{apd:dpb}

%\subsubsection{Efficient Implementation of DPB} \label{apd:dpb-1}

\begin{wrapfigure}[]{r}{0.3\linewidth}
    \vspace{-3mm}
    \includegraphics[width=1.0\linewidth]{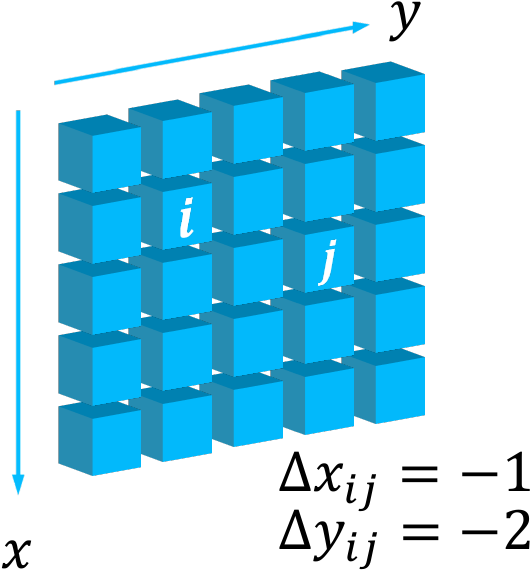}
    \caption{An example of computing $(\Delta x_{ij}, \Delta y_{ij})$.}
    \label{fig:effi_DPB}
    \vspace{-3mm}
\end{wrapfigure}
Figure~\ref{fig:effi_DPB} gives an example of computing $(\Delta x_{ij}, \Delta y_{ij})$ with $G=5$ in the DPB module. For a group of size $G \times G$, it is easy to deduce that:
\begin{equation}
\label{equ:effi_DPB}
\begin{aligned}
0 &\le x, y < G \\
1 - G &\le \Delta x_{ij} \le G - 1 \\
1 - G &\le \Delta y_{ij} \le G - 1.
\end{aligned}
\end{equation}
Thus, motivated by the relative position bias, we construct a matrix $\hat{\mB} \in \mathbb{R}^{(2G-1) \times (2G-1)}$, where
\begin{equation}
\label{equ:effi_DPB-2}
\begin{aligned}
\hat{\mB}_{i, j} = DPB(1-G+i, 1-G+j),\ 0 \le i, j < 2G-1.
\end{aligned}
\end{equation}
The complexity of computing $\hat{\mB}$ is $O(G^2)$. Then, the bias matrix $\mB$ in DPB can be drawn from $\hat{\mB}$, \ie,
\begin{equation}
\label{equ:effi_DPB-3}
\begin{aligned}
\mB_{i, j} = \hat{\mB}_{\Delta x_{ij}, \Delta y_{ij}}.
\end{aligned}
\end{equation}
When the image/group size (\ie, $G$) is fixed, both $\hat{\mB}$ and $\mB$ will be also unchanged in the test phase. Therefore, we only need to compute $\hat{\mB}$ and $\mB$ once, and DPB is equivalent to relative position bias in this case.

\subsection{Variants of CrossFormer for Detection and Segmentation} \label{apd:variants}
\begin{table}[]
    \centering
    \caption{CrossFormer-based backbones for object detection and semantic/instance segmentation. The example input size is $1280 \times 800$. $D$ and $H$ mean embedding dimension and the number of heads in the multi-head self-attention module, respectively. $G$ and $I$ are group size and interval for SDA and LDA, respectively.}
    \scalebox{0.7}{
        \setlength{\tabcolsep}{3mm}{
            \begin{tabular}{cc|c|cccc}
                \toprule
                & Output Size & Layer Name & CrossFormer-T & CrossFormer-S & CrossFormer-B & CrossFormer-L \\ \midrule
                \multirow{5}{*}{Stage-1} & \multirow{5}{*}{$ 320 \times 200 $} & Cross Embed. & \multicolumn{4}{c}{Kernel size: $4 \times 4$, $8 \times 8$, $16 \times 16$, $32 \times 32$, Stride=$4$} \\
                \cmidrule{3-7} 
                & & \multirow{3}{*}{SDA/LDA} & \multirow{2}{*}{$\begin{bmatrix}D_1=64 \\H_1=2 \\G_1=14 \\I_1=16\end{bmatrix} \times 1$} & \multirow{2}{*}{$\begin{bmatrix}D_1=96 \\H_1=3 \\G_1=14 \\I_1=16\end{bmatrix} \times 2$} & \multirow{2}{*}{$\begin{bmatrix}D_1=96\\H_1=3\\G_1=14\\I_1=16, \end{bmatrix} \times 2$} & \multirow{2}{*}{$\begin{bmatrix}D_1=128 \\ H_1=4 \\ G_1=14 \\ I_1=16\end{bmatrix} \times 2$} \\
                &  & \multirow{3}{*}{MLP} &  &  &  \\
                &  &  & & & & \\
                &  &  & & & & \\
                \midrule
                \multirow{5}{*}{Stage-2} & \multirow{5}{*}{$ 160\times100 $} & Cross Embed. & \multicolumn{4}{c}{Kernel size: $2 \times 2$, $4 \times 4$, Stride=$2$} \\ 
                \cmidrule{3-7} 
                & & \multirow{3}{*}{SDA/LDA} & \multirow{3}{*}{$\begin{bmatrix}D_2=128\\H_2=4 \\ G_2=14\\I_2= 8 \end{bmatrix} \times 1$} & \multirow{2}{*}{$\begin{bmatrix}D_2=192\\H_2=6 \\ G_2=14\\I_2= 8 \end{bmatrix} \times 2$} & \multirow{2}{*}{$\begin{bmatrix}D_2=192\\H_2=6 \\ G_2=14\\I_2=8 \end{bmatrix} \times 2$} & \multirow{2}{*}{$\begin{bmatrix}D_2=256\\H_2=8 \\ G_2=14\\I_2=8 \end{bmatrix} \times 2$} \\
                &  & \multirow{3}{*}{MLP} &  &  &  \\
                &  &  & & & & \\
                &  &  & & & & \\
                \midrule
                \multirow{5}{*}{Stage-3} & \multirow{5}{*}{$ 80\times50 $} & Cross Embed. & \multicolumn{4}{c}{Kernel size: $2 \times 2$, $4 \times 4$, Stride=$2$} \\ 
                \cmidrule{3-7} 
                & & \multirow{3}{*}{SDA/LDA} & \multirow{3}{*}{$\begin{bmatrix}D_3=256\\H_3=8 \\ G_3=7\\I_3=2 \end{bmatrix} \times 8$} & \multirow{2}{*}{$\begin{bmatrix}D_3=384\\H_3=12 \\ G_3=7\\I_3=2 \end{bmatrix} \times 6$} & \multirow{2}{*}{$\begin{bmatrix}D_3=384\\H_3=12 \\ G_3=7\\I_3=2 \end{bmatrix} \times 18$} & \multirow{2}{*}{$\begin{bmatrix}D_3=512\\H_3=16 \\ G_3=7\\I_3=2 \end{bmatrix} \times 18$} \\
                &  & \multirow{3}{*}{MLP} &  &  &  \\ 
                &  &  & & & & \\
                &  &  & & & & \\
                \midrule
                \multirow{5}{*}{Stage-4} & \multirow{5}{*}{$ 40 \times 25 $} & Cross Embed. & \multicolumn{4}{c}{Kernel size: $2 \times 2$, $4 \times 4$, Stride=$2$} \\ 
                \cmidrule{3-7}
                & & \multirow{3}{*}{SDA/LDA} & \multirow{3}{*}{$\begin{bmatrix}D_4=512\\H_4=16 \\ G_4=7\\I_4=1 \end{bmatrix} \times 6$} & \multirow{2}{*}{$\begin{bmatrix}D_4=768\\H_4=24 \\ G_4=7\\I_4=1 \end{bmatrix} \times 2$} & \multirow{2}{*}{$\begin{bmatrix}D_4=768\\H_4=24 \\ G_4=7\\I_4=1 \end{bmatrix} \times 2$} & \multirow{2}{*}{$\begin{bmatrix}D_4=1024\\H_4=32 \\ G_4=7\\I_4=1 \end{bmatrix} \times 2$} \\
                &  & \multirow{3}{*}{MLP} &  &  &  \\ 
                &  &  & & & & \\
                &  &  & & & & \\
                \bottomrule
    \end{tabular}}}
    \label{tab:variants-2}
\end{table}
We test two different backbones for dense prediction tasks. The variants of CrossFormer for  dense prediction (object detection, instance segmentation, and semantic segmentation) are in Table~\ref{tab:variants-2}. The architectures are the same as those for image classification except that different $G$ and $I$ in the first two stages are used. Notably, group size (\ie, $G$ and $I$) does not affect the shape of weight tensors, so backbones pre-trained on ImageNet can be fine-tuned directly on other tasks even if they use different $G$ and $I$.
    
\section{Experiments}

\begin{table}[t]
    \centering
    \caption{Object detection results on COCO \textit{val} 2017. ``Memory'' means the allocated memory per GPU reported by $torch.cuda.max\_memory\_allocated()$. $^\ddagger$ indicates that models use different $(G, I)$ from classification models.}
    \scalebox{0.8}{
        \begin{tabular}{c|l|ccccr|rr|lll}
            \toprule
            Method & Backbone & $G_1$ & $I_1$ & $G_2$ & $I_2$ & Memory & \#Params & FLOPs & AP$^b$ & AP$^b_{50}$ & AP$^b_{75}$   \\
            \midrule
            \multirow{4}{*}{RetinaNet} & CrossFormer-S & 7 & 8 & 7 & 4 & 14.7G & 40.8M & 282.0G & 44.4 & 65.8 & 47.4\\
            \multirow{4}{*}{$1\times$ schedule} & CrossFormer-S$^\ddagger$& 14 & 16 & 14 & 8 & 11.9G & 40.8M & 272.1G & 44.2 & 65.7 & 47.2\\
            \cmidrule{2-12}
            & CrossFormer-B & 7 & 8 & 7 & 4 & 22.8G & 62.1M & 389.0G & 46.2 & 67.8 & 49.5 \\
            & CrossFormer-B$^\ddagger$& 14 & 16 & 14 & 8 & 20.2G & 62.1M & 379.0G & 46.1 & 67.7 & 49.0 \\
            \midrule
            \multirow{4}{*}{Mask-RCNN} & CrossFormer-S & 7 & 8 & 7 & 4 & 15.5G & 50.2M & 301.0G & 45.4 & 68.0 & 49.7 \\
            \multirow{4}{*}{$1\times$ schedule} & CrossFormer-S$^\ddagger$& 14 & 16 & 14 & 8 & 12.7G & 50.2M & 291.1G & 45.0 & 67.9 & 49.1 \\
            \cmidrule{2-12}
            & CrossFormer-B & 7 & 8 & 7 & 4 &23.8G & 71.5M & 407.9G & 47.2 & 69.9 & 51.8  \\
            & CrossFormer-B$^\ddagger$& 14 & 16 & 14 & 8 & 21.0G & 71.5M & 398.1G & 47.1 & 69.9 & 52.0 \\
            \bottomrule
    \end{tabular}}
    \label{tab:apd-detection}
\end{table}

\begin{table}[t!]
    \caption{Semantic segmentation results on ADE20K validation set with semantic FPN or UPerNet as heads.}
    \centering
    \scalebox{0.72}{
        \begin{tabular}{l|cccc|rrrr|rrrrr}
            \toprule
            \multirow{2}{*}{Backbone} & \multirow{2}{*}{$G_1$} & \multirow{2}{*}{$I_1$} & \multirow{2}{*}{$G_2$} & \multirow{2}{*}{$I_2$} & \multicolumn{4}{c|}{Semantic FPN ($80$K iterations)} & \multicolumn{5}{c}{UPerNet ($160$K iterations)} \\
            & & & & & Memory & \#Params & FLOPs & IOU & Memory & \#Params & FLOP & IOU & MS IOU \\
            \midrule
            CrossFormer-S & 7 & 8 & 7 & 4 & 20.9G & 34.3M & 220.7G & 46.0 & $-$ & 62.3M & 979.5G & 47.6 & 48.4 \\
            CrossFormer-S$^\ddagger$& 14 & 16 & 14 & 8 & 20.9G & 34.3M & 209.8G & 46.4 & 14.6G & 62.3M & 968.5G & 47.4 & 48.2 \\
            \midrule
            CrossFormer-B & 7 & 8 & 7 & 4 & 14.6G & 55.6M & 331.0G & 47.7 & 15.8G & 83.6M & 1089.7G & 49.7 & 50.6 \\
            CrossFormer-B$^\ddagger$& 14 & 16 & 14 & 8 & 14.6G & 55.6M & 320.1G & 48.0 & 15.8G & 83.6M & 1078.8G & 49.2 & 50.1 \\
            \midrule
            CrossFormer-L & 7 & 8 & 7 & 4 & 25.3G & 95.4M & 497.0G & 48.7 & 18.1G & 125.5M & 1257.8G & 50.4 & 51.4 \\
            CrossFormer-L$^\ddagger$& 14 & 16 & 14 & 8 & 25.3G & 95.4M &  482.7G & 49.1 & 18.1G & 125.5M & 1243.5G & 50.5 & 51.4 \\
            \bottomrule
    \end{tabular}}
    %   \end{subtable}
    \label{tab:apd-segmentation-1}
\end{table}

\subsection{Object Detection}

Table~\ref{tab:apd-detection} provides more results on object detection with RetinaNet and Mask-RCNN as detection heads. As we can see, a smaller $(G, I)$ achieves a higher AP than a larger one, but the performance gain is marginal. Considering that a larger $(G, I)$ can save more memory cost, we think $(G_1=14, I_1=16, G_2=14, I_2=8)$, which accords with configurations in Table~\ref{tab:variants-2}, achieves a better trade-off between the performance and cost.

\subsection{Semantic Segmentation}

Similar to object detection, we test two different configurations of $(G, I)$ for semantic segmentation's backbones. The results are shown in Table~\ref{tab:apd-segmentation-1}. As we can see, the memory costs of the two configurations are almost the same, which is different from experiments on object detection. Further, when taking semantic FPN as the detection head, CrossFormers$^\ddagger$ show advantages over CrossFormers on both IOU (\eg, $46.4$ vs. $46.0$) and FLOPs (\eg, $209.8$G vs. $220.7$G). When taking UPerNet as the segmentation head, a smaller $(G, I)$ achieves higher performance like object detection.
    
\end{document}